\documentclass{article} 
\usepackage{iclr2026_conference,times}
\usepackage{graphicx}

\usepackage{amsmath,amsfonts,bm}

\def\eqref#1{equation~\ref{#1}}

\def\1{\bm{1}}

\DeclareMathAlphabet{\mathsfit}{\encodingdefault}{\sfdefault}{m}{sl}
\SetMathAlphabet{\mathsfit}{bold}{\encodingdefault}{\sfdefault}{bx}{n}

\usepackage{hyperref}
\usepackage{url}
\usepackage{amsmath}
\usepackage{amssymb}
\usepackage{mathtools}
\usepackage{bm}
\usepackage{caption}
\usepackage{booktabs}
\usepackage{makecell}

\title{Closing the Loop: Training-Free Revisit Consistency for Autoregressive Generative Rendering}

\author{Wenchao Ma$^{1,2*}$, Changran Liu$^{1*}$, Sharon X.\ Huang$^{2}$ \& Haomiao Jiang$^{1}$ \\
$^{1}$Roblox \qquad $^{2}$The Pennsylvania State University \\
$^{*}$Equal contribution
}
\iclrfinalcopy 
\begin{document}

\maketitle
\begin{center}
\vspace{-6mm}
    \captionsetup{type=figure}
    \includegraphics[width=0.8\linewidth]{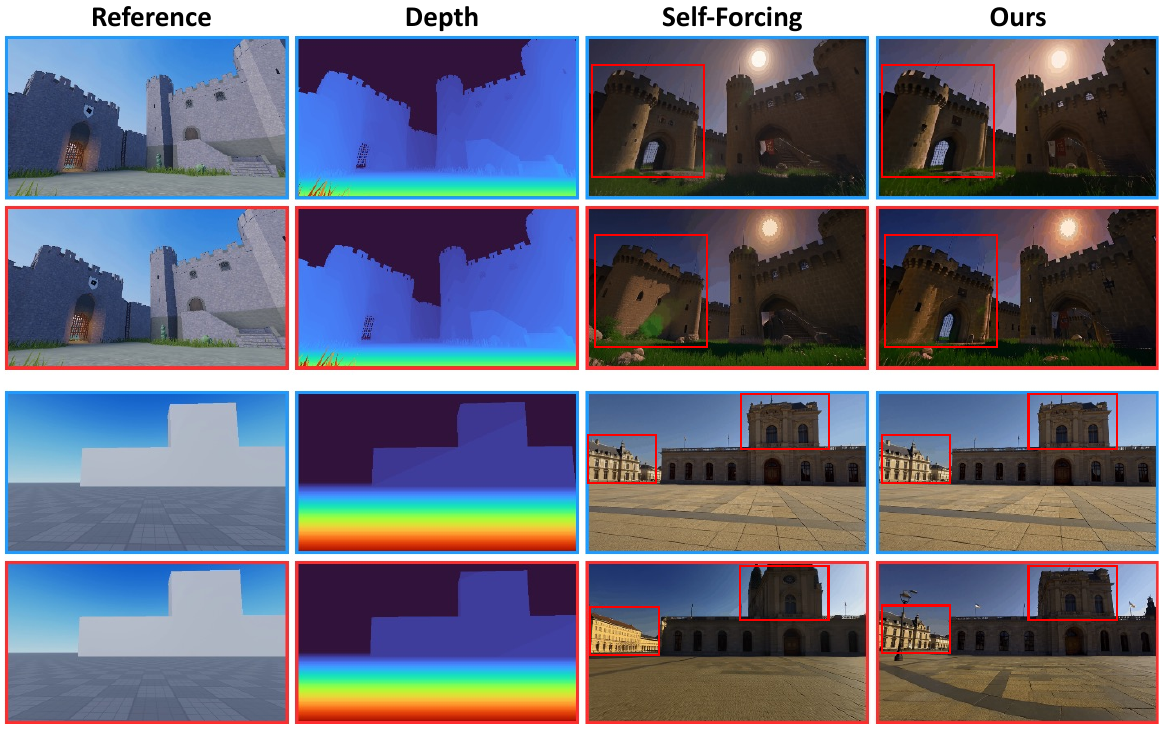}
    \captionof{figure}{\textbf{Revisit-consistent auto-regressive generative rendering.} Each pair of rows shows a revisit clip: conditioned on depth  streamed from a 3D engine (Reference and Depth columns), an autoregressive video generator renders a viewpoint at the first visit (blue) and again when the camera returns (red), far beyond its bounded KV-cache horizon. Our method re-renders the same structures on return.}
    \label{fig:teaser}
\end{center}

\begin{abstract}
Recent conditional video generation models have shown promising potentials to transform 3D engine renderings, such as depth maps and untextured geometry, into photorealistic videos for gaming and immersive content creation. These applications require long-horizon auto-regressive generation that continuously synthesizes new frames while preserving a persistent 3D world. Auto-regressive generators synthesize video chunk by chunk with a bounded KV cache, so when the camera revisits a location after its context has been evicted, the model often regenerates inconsistent appearance, even though the conditioning renderings (e.g., depth) remain perfectly aligned with the underlying geometry.We address this revisit inconsistency without any post-training by exploiting correspondences the 3D engine already provides: temporal correspondence retrieves pose-matched historical latent chunks into the KV cache as loop-closure memory, while spatial correspondence from camera pose and depth reprojection biases token-level attention toward geometrically corresponding regions of the retrieved chunks. We demonstrate our method on loop-closure trajectories mined from TartanAir and TartanGround dataset to mirror complicate real-world application scenarios, where it outperforms existing training-free baselines on revisit consistency without losing overall video quality. Project Page: \href{https://wenchao-m.github.io/ClosetheLoop.github.io/}{\textcolor{pink}{https://wenchao-m.github.io/ClosetheLoop.github.io/}}
\end{abstract}

~\section{Introduction}
Generative rendering aims to lift low-fidelity yet precisely controllable outputs from 3D engines into realistic visual content. Given engine-provided conditions such as depth maps, untextured geometry, semantic buffers, or coarse simulations, a generative model can synthesize photorealistic images or videos while preserving the controllability of the underlying graphics pipeline. This capability is especially attractive for gaming, virtual production, simulation, and immersive content creation, where users require precise control over camera motion, scene layout, object trajectories, and interactions, while also seeking visual realism that would otherwise demand costly asset authoring, material design, lighting, and rendering in traditional graphics pipelines.

Prior arts~\cite{wang2018vid2vid,mallya2020world,cai2024generative,gomez2026coarse,cohenbar2026realmaster,abualhaija2025cosmostransfer,nvidia2025worldsimulation,jiang2025vace,fal2026ltx23render2real} have explored this direction by combining 3D or simulator-derived conditions with image- or video-generation models~\cite{isola2017pix2pix, wang2018pix2pixhd,rombach2022latent,peebles2023dit,zhang2023controlnet,wang2025wan,hacohen2025ltxvideo}. These methods demonstrate the promise of generative rendering, but they are primarily designed for short or offline video clips, where the necessary context remains within the image/video generation window or can be propagated through an offline pipeline.

\begin{figure}
    \centering
    \includegraphics[width=0.9\linewidth]{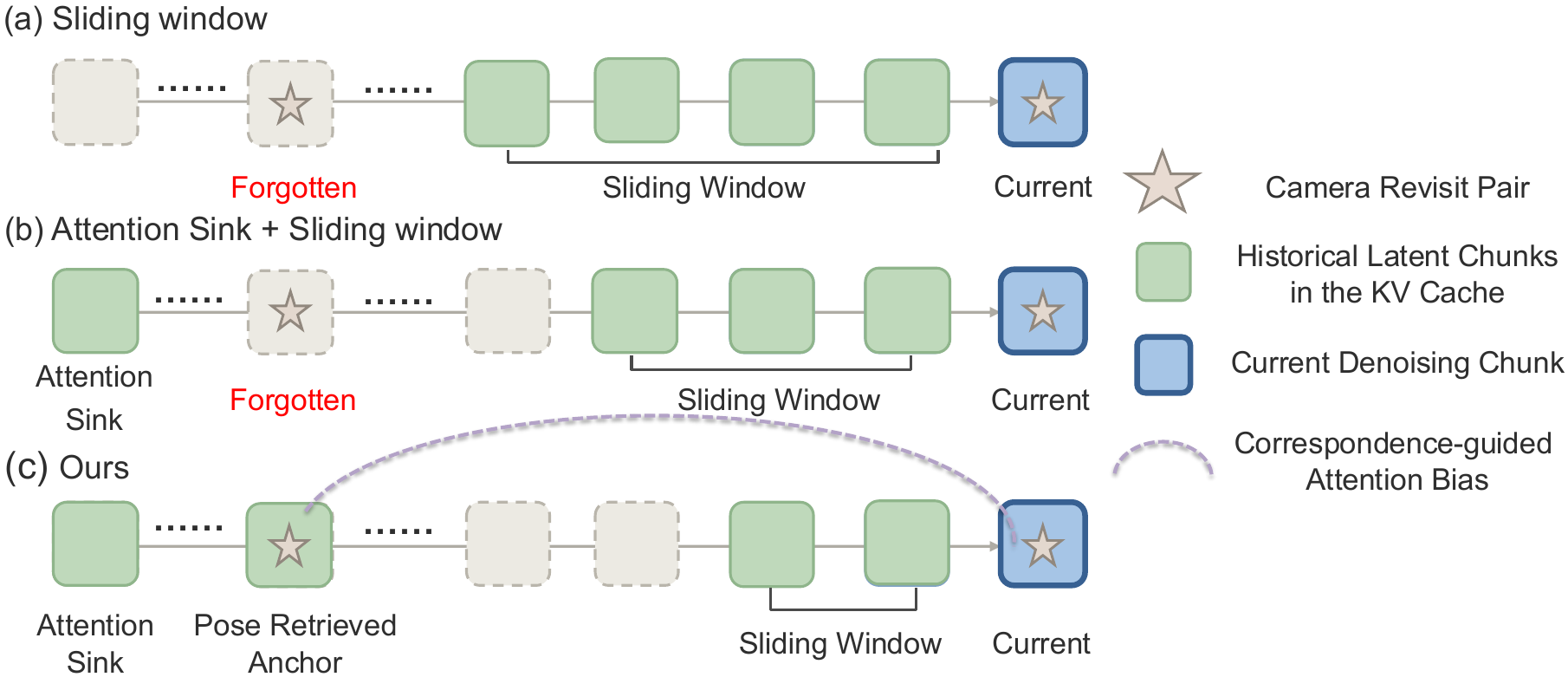}
    \caption{KV cache strategies under camera revisits. Naive sliding-window caching (a), or with an attention sink (b), forgets chunks from previously visited viewpoints. Ours (c) retrieves the pose-matched historical chunk as an anchor in the KV cache and applies a spatial correspondence-guided attention bias that links current-frame tokens to corresponding regions in the retrieved frame.}
    \label{fig:kv_compare}
\vspace{-2mm}
\end{figure}

However, many practical applications demand long-horizon streaming generation. In an interactive game or virtual environment, for instance, the camera moves continuously through a large scene, leaves a region, and revisits it much later; the generator must synthesize frames chunk by chunk while maintaining a persistent world over horizons far exceeding the native context window of the base video model. Recent distillation methods for autoregressive video diffusion make such streaming feasible through causal attention and a bounded KV cache~\cite{chen2024diffusionforcing,yin2025causvid,huang2025selfforcing,yang2025longlive,zhu2026causalforcing,zhao2026causalforcingpp}. Yet bounded memory introduces a new failure mode for generative rendering, illustrated in Fig.~\ref{fig:kv_compare}: under a plain sliding window (Fig.~\ref{fig:kv_compare}a), the chunk depicting a revisited location has long been evicted by the time the camera returns. Motivated by StreamingLLM~\cite{XiaoTCHL24}, some works~\cite{shin2025motionstream, yesiltepe2025infinityrope} add a fixed attention sink (Fig.~\ref{fig:kv_compare}b) that preserves the first chunk, mitigating color drift and improving consistency when the foreground remains unchanged throughout the video. But the first chunk in general does not coincide with the revisited view, so this sink still cannot anchor a loop closure. In both cases the model must regenerate the location from scratch and may produce different textures, colors, or structures, even though the conditioning depth and camera trajectory are perfectly consistent since they originate from the same underlying 3D geometry.

We refer to this failure as \emph{revisit inconsistency}. Prior work on world-consistent video generation combats forgetting by maintaining explicit 3D state, geometric view memories, hierarchical latents, or learned context-querying and memory modules that can be reused when content reappears~\cite{mallya2020world,ren2025gen3c,garcin2026persist,li2025vmem,chandratreya2026millivid,chen2026hybridmemory,yu2026memlearner}. These approaches underscore that memory beyond a local history is essential, but they typically depend on costly curated long-horizon datasets and additional model training. In this work, we ask: \textit{how can we leverage the temporal and spatial correspondences already available from a 3D engine to improve revisit consistency in a pretrained autoregressive video generator, without expensive video-model training?}

To this end, we propose a correspondence-guided memory framework for long-horizon auto-regressive generative rendering (Fig~\ref{fig:kv_compare} (c)). Temporally, we retrieve historical latent chunks whose camera poses match the current chunk and reintroduce them into the clean KV cache as loop-closure memory. This mechanism gives the model access to the actual previously generated view, extending its usable memory beyond the default cache window. Spatially, we use camera pose and metric depth to reproject each current query token into the selected historical memory. Rather than directly warping and blending historical features, which can introduce blur and artifacts, we convert the correspondence into a Gaussian bias on the historical-memory attention logits. This encourages each current token to attend to geometrically corresponding historical tokens to improve the revisit consistency. 

Our experiments show that the proposed correspondence-guided memory substantially improves revisit consistency in streaming generative rendering. On a loop-closure benchmark we construct from TartanGround~\cite{patel2025tartanground} and TartanAir~\cite{wang2020tartanair}---photorealistic navigation trajectories that repeatedly leave and return to the same locations---our method consistently outperforms existing training-free cache-management schemes on the pixel, semantic, and geometric-correspondence metrics of MilliVid~\cite{chandratreya2026millivid}, without degrading overall video quality as measured by VBench-Long~\cite{huang2025vbench++}. We also showcase integrating our method with a live game engine, demonstrating revisit-consistent generative rendering directly from engine-streamed depth and camera poses.

~\section{Related Work}

\subsection{Generative Rendering}

Generative rendering combines the controllability of graphics engines with the realism of generative models. Early video-to-video synthesis methods translate structured conditions, such as semantic layouts or rendered guidance, into photorealistic videos using conditional GANs~\cite{wang2018vid2vid,mallya2020world}. Recent diffusion-based methods improve visual quality by conditioning image or video diffusion model on 3D or simulator-derived signals. Generative Rendering~\cite{cai2024generative} uses dynamic mesh correspondences with a pretrained 2D diffusion model to synthesize temporally consistent frames from low-fidelity animated mesh renderings. Coarse-to-Real~\cite{gomez2026coarse} generates realistic populated urban videos from coarse 3D simulations, while RealMaster~\cite{cohenbar2026realmaster} lifts 3D-engine rendered videos into photorealistic videos while preserving renderer-specified geometry and dynamics. Large conditional video models such as Cosmos-Transfer and WanVACE further demonstrate controllable video generation from spatial controls that could be extracted from graphics engine, including depth, segmentation, edges, and rendered videos~\cite{abualhaija2025cosmostransfer,nvidia2025worldsimulation,wang2025wan,jiang2025vace}. These methods mainly target short or offline clips, whereas we study long-horizon auto-regressive generative rendering with bounded KV memory and revisit inconsistency.

\subsection{Consistent Video Generation}

Long-term consistency has been studied in video world models through explicit memory, retrieval, and compact historical state. Seoul World Model~\cite{seo2026seoulworldmodel} grounds world simulation in a real metropolis by retrieving relevant street-view context from a large database. VMem~\cite{li2025vmem} maintains a surfel-indexed view memory, enabling previously generated views to be stored and reused in an interactive scene. Matrix-Game 3.0~\cite{wang2026matrixgame3} uses camera-pose-based memory buffers for real-time streaming world modeling. Hybrid Memory~\cite{chen2026hybridmemory} separates memory for static backgrounds and dynamic objects to preserve content after it leaves the visible field. RELIC~\cite{hong2025relic} introduces compact long-horizon memory for interactive video world models, while MemLearner~\cite{yu2026memlearner} learns to query useful context memory adaptively. MilliVid~\cite{chandratreya2026millivid} uses hierarchical latents for long-range consistency, and PERSIST~\cite{garcin2026persist} maintains a persistent 3D state rather than relying only on pixel histories. These approaches highlight the importance of memory beyond local context, but generally require  curated long-horizon data and costly model pretraining.

Another related line improves long-video generation through post-training, distillation, or inference-time memory design. Memorize-and-Generate~\cite{zhu2025mag} trains a memory block to compress historical information for real-time generation. MotionStream~\cite{shin2025motionstream} combines self-forcing, sliding-window causal attention, and attention sinks for interactive motion-controlled video generation. VideoSSM~\cite{yu2025videossm} augments autoregressive generation with hybrid state-space memory, while Helios~\cite{yuan2026helios} uses compressed historical context in a multi-stage real-time generation pipeline. Deep Forcing~\cite{yi2025deepforcing} proposes training-free KV-cache management with deep sinks and participative compression, Infinity-RoPE~\cite{yesiltepe2025infinityrope} modifies RoPE and cache handling for autoregressive self-rollout, and LongLive-RAG~\cite{hu2026longliverag} retrieves historical latents for long video generation. MemRoPE~\citep{kim2026memrope} maintains a fixed set of evolving memory tokens that continuously compress past keys via exponential moving averages. Our work is complementary to these approaches. Instead of learning or compressing generic video history, we exploit correspondences already available from the 3D engine: temporal correspondence retrieves loop-closure chunks into the KV cache, while spatial correspondence biases attention toward geometrically matched historical tokens.

~\section{Preliminaries}
\label{sec:preliminaries}

\subsection{Auto-regressive Video Generation with Self-Forcing}
\label{sec:prelim_sf}
Autoregressive video diffusion models generate long videos in latent chunks: a causal generator denoises the current chunk while attending to a fixed-size sliding window of previously generated chunks through a key--value (KV) cache, so memory and per-chunk compute remain constant as the video grows. Self-Forcing~\cite{huang2025selfforcing} distills a bidirectional video diffusion model teacher into such a causal, few-step student by making training mirror this inference process. Instead of conditioning on ground-truth history, the student rolls out from pure Gaussian noise; after each chunk is denoised, a clean-context pass writes its keys and values into a clean KV cache that later chunks attend to, exposing the student to its own accumulated errors and reducing the train--test mismatch of teacher forcing. Supervision follows distribution-matching distillation~\cite{yin2024dmd,yin2024dmd2}: a frozen teacher provides the target score, a jointly trained fake-score network estimates the score of the student's samples, and their difference drives the student update, optionally with a GAN discriminator on teacher features.

\subsection{Causal Distillation of Depth-Conditioned Wan-VACE}
\label{sec:prelim_vace}

We instantiate this recipe on Wan-VACE~\cite{jiang2025vace}, which conditions the Wan text-to-video (T2V) DiT~\cite{wang2025wan} on depth: normalized monocular depth maps are encoded by the Wan VAE and processed by VACE control blocks, whose outputs are injected as residuals into selected backbone layers. Since the pretrained model is bidirectional, we causalize both the main branch and the VACE control branch, each with its own KV cache, to obtain Causal Wan-VACE. Training then follows Sec.~\ref{sec:prelim_sf}: the teacher and fake-score networks remain bidirectional Wan-VACE, and a single depth-control context is shared by all networks. We use this self-forced, depth-conditioned Causal Wan-VACE as our baseline model.

\section{Method}
\label{sec:method}


Our goal is long-horizon revisit consistency: when the camera leaves a region and returns after the corresponding chunks have been evicted from the clean KV cache, the regenerated content should be consistent with what was generated before. We address this with two complementary, purely inference-time mechanisms that operate on the frozen casual student of Sec.~\ref{sec:prelim_vace}. First, \emph{pose-retrieved loop-closure memory} (Sec.~\ref{sec:method_memory}) decides \emph{what} the model can see: it detects that the current camera pose revisits a previously generated viewpoint and reinstates the corresponding historical latent chunk into the clean cache. Second, a \emph{geometry-guided attention bias} (Sec.~\ref{sec:method_bias}) decides \emph{where} the model looks: using the known correspondence from the 3D engine, each current query token is softly steered toward its spatial corresponding token in the retrieved chunks.

\subsection{Setup and Assumptions}
\label{sec:method_setup}

The student generates latent chunks $\mathbf{x}^{(1)}, \mathbf{x}^{(2)}, \ldots$ of $F$ latent frames each ($F{=}3$ in our implementation), and self-attention operates on an $h \times w$ token grid per latent frame ($30 \times 52$ for $832 \times 480$ outputs in Wan-VACE). We assume that each latent frame $f$ is annotated with pinhole intrinsics $\mathbf{K}_f$, rescaled to the token grid, and a rigid camera pose, written as a camera-to-world rotation and center $(\mathbf{R}_f, \mathbf{c}_f)$. We also assume per-frame planar metric depth $D_f$, i.e., distance along the optical axis in scene units consistent with the camera translations. Both annotations are natural byproducts of the generative-rendering setting we target, where a 3D engine supplies the depth-control video generation model together with the corresponding camera parameters and metric depth.

\subsection{Pose-Retrieved Loop-Closure Memory}
\label{sec:method_memory}
 
\begin{figure}
    \centering
    \includegraphics[width=0.9\linewidth]{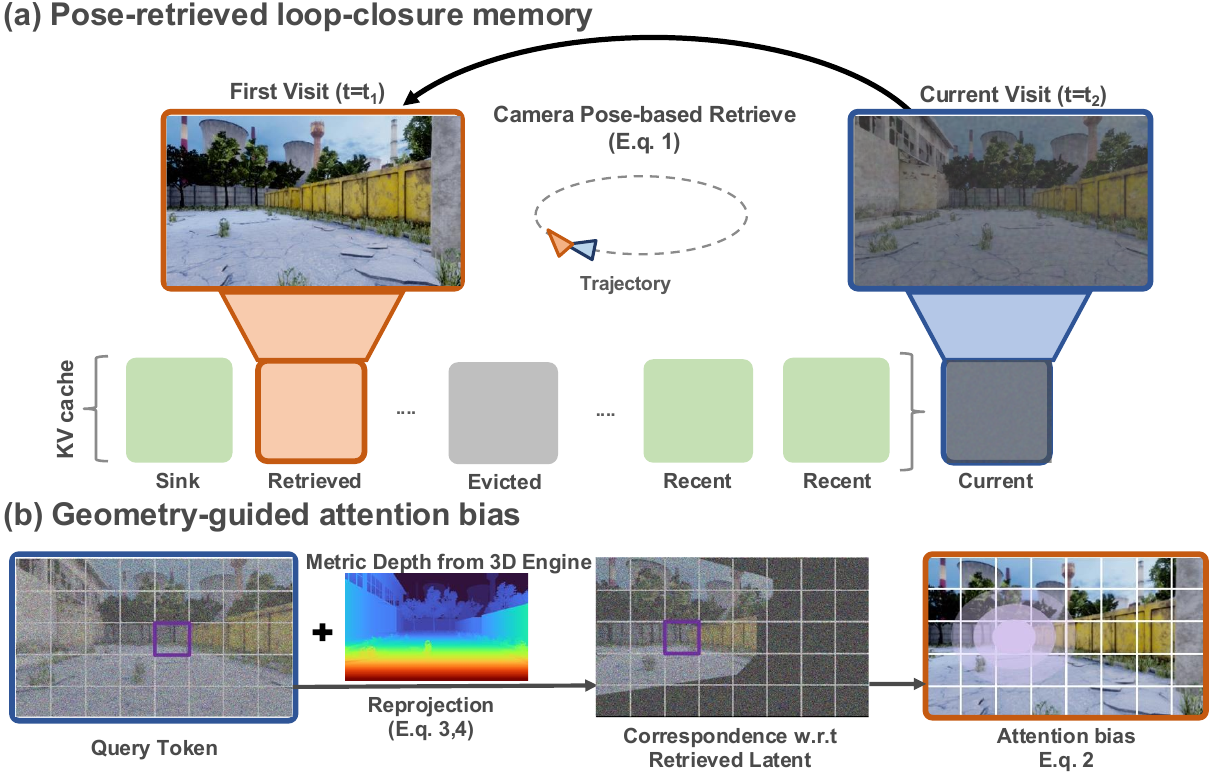}
    \caption{
\textbf{(a)}~When the current pose ($t{=}t_2$) returns near a previously generated viewpoint ($t{=}t_1$), pose-based retrieval (Eq.~\ref{eq:retrieval}) reinstates the evicted chunk into the bounded KV cache.
\textbf{(b)}~Each query token is reprojected into the retrieved view via metric depth and relative pose (Eq.~\ref{eq:reprojection} and Eq.~\ref{eq:visibility}), and a Gaussian bias at the predicted correspondence is added to the attention logits (Eq.~\ref{eq:bias}); no cached features are warped.}
\label{fig:model_detail}
\end{figure}

\paragraph{Cache layout.}
As shown in the Fig~\ref{fig:model_detail} (a), we partition the clean-cache budget of $M$ latent frames into three slots: a persistent \emph{anchor} holding the first chunk, one \emph{pose-retrieved} chunk, and a \emph{recent} window of the latest chunks; we use $M{=}4F{=}12$, with $F$ anchor, $F$ retrieved, and $2F$ recent frames. The anchor provides a fixed global reference, the recent window preserves local temporal continuity, and the retrieved slot supplies view-specific long-term memory. Clean latents and poses of evicted chunks are retained outside the attention cache so that any historical chunk can be retrieved.
 

\paragraph{Loop-closure retrieval.}
Let $(\mathbf{c}_n, \mathbf{v}_n)$ denote the camera center and unit viewing direction associated with the current chunk $n$, and $(\mathbf{c}_j, \mathbf{v}_j)$ those of a historical chunk $j$. We reinstate the admissible candidate closest in pose to the current chunk:
\begin{equation}
\begin{aligned}
s(j) \;&=\; \lVert \mathbf{c}_j - \mathbf{c}_n \rVert_2 / E \;+\; w_v \bigl(1 - \mathbf{v}_j^{\top}\mathbf{v}_n\bigr), \\[2pt]
\mathcal{A}_n \;&=\; \bigl\{\, j \;:\; \lVert \mathbf{c}_j - \mathbf{c}_n \rVert_2 / E \,\le\, \tau_c,\;\; \mathbf{v}_j^{\top}\mathbf{v}_n \,\ge\, \cos\tau_v,\;\; n - j \,\ge\, \Delta \,\bigr\},
\end{aligned}
\label{eq:retrieval}
\end{equation}
where $E$ is a scene-scale normalizer defined by the scene size and $w_v$ (0.5 in our setting) weights the angular term. The admissible set $\mathcal{A}_n$ encodes three gates: a normalized return distance below $\tau_c$, viewing directions within $\tau_v$, and a temporal gap of at least $\Delta$ chunks. If $\mathcal{A}_n$ is empty, the retrieved slot simply extends the recent window. Each gate is functionally necessary: the distance and angle constraints ensure that long-range memory is injected only at genuine revisits, while the temporal gap prevents the immediately preceding chunks already present in the recent window from being trivially selected as spurious loop closures.
 
\paragraph{Cache packing and cross-branch alignment.}
The retrieved chunk is packed at RoPE positions directly adjacent to the recent window rather than at its original temporal index, keeping cache positions within a fixed range under unbounded streaming. Because Causal Wan-VACE maintains separate caches for the main and control branches (Sec.~\ref{sec:prelim_vace}), the corresponding slice of the VACE control cache is copied to the same packed location. 

\subsection{Geometric Correspondence as an Attention Prior}
\label{sec:method_bias}
 
Reinstating a chunk restores access to the earlier appearance, but the model must still establish correspondences between current queries and retrieved keys across a potentially large viewpoint change, a matching problem that the self attention mechanism in the video DiT would otherwise have to solve implicitly. Since the geometrical correspondence between the two views is known from the 3D engine, we inject it explicitly as a soft prior on the attention logits (Fig~\ref{fig:model_detail} (b)).
 
\paragraph{General form.}
Consider a query token $p$ on the current (target) frame $t$ and a key token $k$ on a retrieved (source) frame $s$. Their positions on the $h \times w$ token grid are $\mathbf{u}_p$ and $\mathbf{u}_k$, and their query and key features in a given self-attention head are $\mathbf{q}_p, \mathbf{k}_k \in \mathbb{R}^d$. Suppose a geometric warp predicts the source-frame location $\hat{\mathbf{u}}_s(p)$ at which the content observed at $\mathbf{u}_p$ appeared in frame $s$ (constructed below). We penalize each key by its distance to this predicted correspondence through a Gaussian bias $B_s(p,k)$, added to the standard scaled dot-product logit:
\begin{equation}
B_s(p,k) \;=\; -\,\frac{\lVert \mathbf{u}_k - \hat{\mathbf{u}}_s(p) \rVert_2^2}{2\sigma^2},
\qquad
A_{pk} \;=\; \frac{\mathbf{q}_p^{\top}\mathbf{k}_k}{\sqrt{d}} \;+\; \lambda\, B_s(p,k),
\label{eq:bias}
\end{equation}
where $\sigma$ (in grid units) sets the spatial bandwidth of the prior and $\lambda$ its strength; logits of all other key columns are unchanged. Because $B_s(p,k) \le 0$ and vanishes exactly at the predicted correspondence, the bias acts multiplicatively on the attention weights as a factor $\exp\!\bigl(\lambda B_s(p,k)\bigr) \in (0,1]$: the key at $\hat{\mathbf{u}}_s(p)$ retains its original weight, while keys farther away are exponentially suppressed. The prior is therefore a Gaussian-shaped suppression mask rather than an additive bonus, it narrows \emph{where} the model looks but cannot force content into place, leaving \emph{what} to copy to the learned attention. 
 
\paragraph{Reprojection warp.}
We obtain $\hat{\mathbf{u}}_s(p)$ by lifting the query to 3D with its planar metric depth $z_p = D_t(\mathbf{u}_p)$ and reprojecting it through the camera poses:
\begin{equation}
\mathbf{X}_t \;=\; z_p\, \mathbf{K}_t^{-1} \tilde{\mathbf{u}}_p,
\qquad
\mathbf{X}_s \;=\; \mathbf{R}_s^{\top}\bigl(\mathbf{R}_t\, \mathbf{X}_t + \mathbf{c}_t - \mathbf{c}_s\bigr),
\qquad
\hat{\mathbf{u}}_s(p) \;=\; \pi\!\bigl( \mathbf{K}_s\, \mathbf{X}_s \bigr),
\label{eq:reprojection}
\end{equation}
where $\tilde{\mathbf{u}}_p$ denotes homogeneous grid coordinates and $\pi(\cdot)$ perspective projection: the query is unprojected into the target camera, expressed in world coordinates through the camera-to-world pose, re-expressed in the source camera, and projected. Because each token is displaced according to its own depth, foreground and background move by different amounts under camera translation, as parallax requires. When the camera center is fixed ($\mathbf{c}_s = \mathbf{c}_t$), depth cancels under perspective projection and Eq.~\eqref{eq:reprojection} reduces to the rotation, zoom homography $\pi(\mathbf{K}_s \mathbf{R}_s^{\top} \mathbf{R}_t \mathbf{K}_t^{-1} \tilde{\mathbf{u}}_p)$.
 
\paragraph{Occlusion-aware visibility.}
The reprojected point may nevertheless be occluded in the source view. At disocclusions, a newly exposed background token would otherwise be steered toward the foreground surface that covered it, producing duplicated structure. We therefore accept a correspondence only if it passes a source-depth visibility test,
\begin{equation}
\bigl[\mathbf{X}_s\bigr]_z \;\le\; (1+\varepsilon)\, D_s\!\bigl(\hat{\mathbf{u}}_s(p)\bigr),
\label{eq:visibility}
\end{equation}
where $[\mathbf{X}_s]_z$ is the point's depth in the source camera and $\varepsilon$ an occlusion tolerance; rejected correspondences receive no bias.
 
\paragraph{Design choices.}
Three restrictions define the mechanism. (i)~\emph{Logits only.} We never warp or blend cached keys, values, latents, or hidden states: resampling features at $30 \times 52$ resolution under imperfect poses or depth would inject blur and geometric error directly into the content path, whereas biasing logits keeps the cached values sharp. (ii)~\emph{Retrieved keys only.} The bias is applied in the main-branch self-attention and only to the key columns of the pose-retrieved chunk; keys of the current chunk, the recent window, the anchor, and the entire control branch are untouched, so short-range temporal attention is unaffected. (iii)~\emph{Exact fallback.} Queries whose correspondence is undefined, off-grid, behind the camera, or rejected by~\eqref{eq:visibility}, receive an all-zero bias row and reduce exactly to ordinary attention.
~\section{Experiment}
\subsection{Evaluation Setup}
\label{sec:exp_dataset}
\begin{figure}[htb!]
    \centering
    \includegraphics[width=\linewidth]{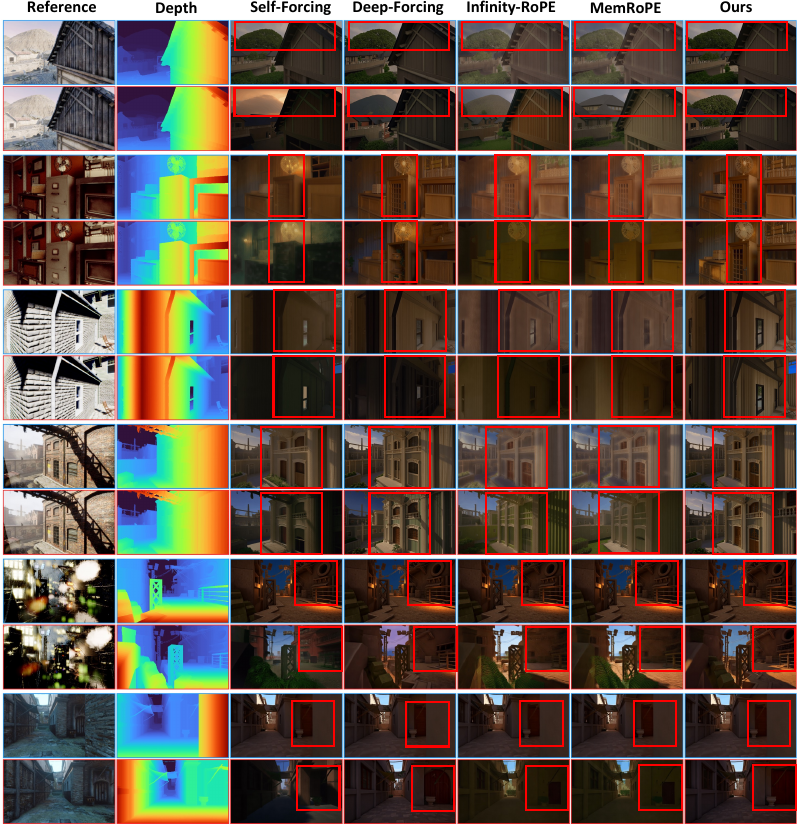}
    \caption{\textbf{Qualitative comparison on loop-closure revisits.} Each pair of rows shows the first visit (blue) and the return (red) of one clip; the Reference and Depth columns show the conditioning source. Red boxes mark the same scene region at both visits. Baselines regenerate altered structures on return; our method re-renders those content consistently.}
\label{fig:qual_compare}
\end{figure}
\paragraph{Evaluation datasets.}
TartanGround~\cite{patel2025tartanground} is a large-scale photorealistic Unreal-Engine dataset built for ground-robot perception and navigation. We repurpose it for revisit evaluation because its navigation trajectories naturally leave and return to places, the engine supplies exactly the annotations of Sec.~\ref{sec:method_setup} (camera poses and planar metric depth), and it is challenging: loops span gaps far beyond any cache horizon, and the translation-dominant motion is parallax-heavy. We mine the loop trajectory from this dataset: a first-visit/return pair qualifies if the camera returns within $2$\,m and $45^{\circ}$ after at least $100$ frames and an excursion of at least $5$\,m (ruling out pauses and in-place turns), yielding $71$ loops (\emph{TartanGround-Revisit}). We additionally construct \emph{TartanAir-Revisit}, a set of controlled revisits built from TartanAir~\cite{wang2020tartanair}. Since the underlying 3D scenes are not publicly released and the original drone trajectories are too jittery for video generation, we cannot re-render new camera paths in the engine; instead, we render smooth revisit trajectories directly from the dataset's $360^{\circ}$ panoramas, scripting two patterns: a \emph{rotation} excursion that yaws away by $90$--$180^{\circ}$ and returns, and a \emph{zoom} excursion that narrows the field of view from $90^{\circ}$ to $35^{\circ}$ and widens back, yielding $285$ clips over $72$ environments.


\paragraph{Evaluation metrics.}
We evaluate both revisit consistency and overall video quality. We followed the evaluation protocol in MilliVid~\cite{chandratreya2026millivid} for consistency evaluation, each loop pair's two generated frames are compared with L1 error and DINOv2~\cite{oquab2023dinov2} cosine similarity. Most diagnostic is geometric correspondence: we detect SuperPoint~\cite{detone2018superpoint} keypoints per frame, match them with LightGlue~\cite{lindenberger2023lightglue}, and count matches with confidence above $0.5$, a high count means structures from the first visit are re-rendered where a matcher can confidently re-localize them, the object permanence loop closure demands. For overall quality, we report VBench-Long~\cite{huang2023vbench,huang2025vbench++,zheng2025vbench2}, verifying that consistency gains do not cost visual fidelity; we omit its dynamic-degree dimension, which saturates at $1.0$ for all methods since motion is controlled by the conditioning depth sequence.

\paragraph{Baseline Methods}
All methods share the same Causal Wan-VACE base model~\citep{jiang2025vace} in Sec~\ref{sec:prelim_vace}, depth control, prompts, random seeds, and KV cache size. We compare against the base model with a plain rolling window (\textbf{Self-forcing}) and three training-free cache-management schemes applied to the same student: \textbf{Infinity-RoPE}~\citep{yesiltepe2025infinityrope}, \textbf{Deep Forcing}~\citep{yi2025deepforcing}, and \textbf{MemRoPE}~\citep{kim2026memrope}.

\begin{table}[t]
\centering
\caption{Loop-closure consistency and overall video quality on the TartanGround benchmark. Consistency metrics are macro-averaged over each clip's genuine loop-closure revisit pairs; quality is measured by VBench-Long (dynamic degree, ${\approx}1.0$ for all methods, is omitted; the mean is over the five reported dimensions). Best in \textbf{bold}. $\uparrow$/$\downarrow$ indicate higher/lower is better.}
\label{tab:tg-revisit}
\resizebox{\linewidth}{!}{%
\begin{tabular}{lccccccccc}
\toprule
& \multicolumn{3}{c}{Loop-closure consistency} & \multicolumn{6}{c}{Video quality (VBench-Long)} \\
\cmidrule(lr){2-4} \cmidrule(lr){5-10}
Method & \makecell{Keypoint\\matches $\uparrow$} & \makecell{DINO\\similarity $\uparrow$} & \makecell{L1\\error $\downarrow$} & \makecell{Subject\\consistency $\uparrow$} & \makecell{Background\\consistency $\uparrow$} & \makecell{Motion\\smoothness $\uparrow$} & \makecell{Aesthetic\\quality $\uparrow$} & \makecell{Imaging\\quality $\uparrow$} & \makecell{Mean $\uparrow$} \\
\midrule
Self Forcing~\cite{huang2025selfforcing} & 23.26 & 0.6087 & 0.1148 & 0.8472 & 0.9371 & \textbf{0.9781} & 0.4627 & 0.3952 & 0.7241 \\
Infinity-RoPE~\cite{yesiltepe2025infinityrope} & 26.89 & 0.6235 & 0.1103 & 0.8239 & 0.9302 & 0.9745 & 0.4542 & 0.4073 & 0.7180 \\
MemRoPE~\cite{kim2026memrope} & 27.07 & 0.6244 & 0.1089 & 0.8227 & 0.9297 & 0.9751 & 0.4537 & 0.3991 & 0.7161 \\
Deep Forcing~\cite{yi2025deepforcing} & 43.90 & 0.6848 & 0.1063 & 0.8586 & 0.9421 & 0.9754 & 0.4738 & 0.4572 & 0.7414 \\
\textbf{Ours} & \textbf{49.08} & \textbf{0.6904} & \textbf{0.1052} & \textbf{0.8592} & \textbf{0.9428} & 0.9754 & \textbf{0.4740} & \textbf{0.4586} & \textbf{0.7420} \\
\bottomrule
\end{tabular}}
\end{table}

\begin{table}[t]\centering
\caption{Loop-closure consistency and overall video quality on the TartanAir revisit benchmark.}
\label{tab:ta-revisit}
\resizebox{\linewidth}{!}{%
\begin{tabular}{lccccccccc}
\toprule
& \multicolumn{3}{c}{Loop-closure consistency} & \multicolumn{6}{c}{Video quality (VBench-Long)} \\
\cmidrule(lr){2-4} \cmidrule(lr){5-10}
Method & \makecell{Keypoint\\matches $\uparrow$} & \makecell{DINO\\similarity $\uparrow$} & \makecell{L1\\error $\downarrow$} & \makecell{Subject\\consistency $\uparrow$} & \makecell{Background\\consistency $\uparrow$} & \makecell{Motion\\smoothness $\uparrow$} & \makecell{Aesthetic\\quality $\uparrow$} & \makecell{Imaging\\quality $\uparrow$} & \makecell{Mean $\uparrow$} \\
\midrule
Self Forcing~\cite{huang2025selfforcing} & 195.65 & 0.7623 & 0.0862 & 0.9163 & 0.9402 & 0.9894 & 0.4592 & 0.4768 & 0.7564 \\
Infinity-RoPE~\cite{yesiltepe2025infinityrope} & 204.14 & 0.7622 & 0.0792 & 0.8901 & 0.9360 & 0.9868 & 0.4446 & 0.4487 & 0.7412 \\
MemRoPE~\cite{kim2026memrope} & 207.95 & 0.7637 & 0.0769 & 0.8882 & 0.9357 & 0.9867 & 0.4428 & 0.4478 & 0.7402 \\
Deep Forcing~\cite{yi2025deepforcing} & 267.87 & 0.8291 & 0.0664 & 0.9271 & \textbf{0.9475} & \textbf{0.9897} & 0.4682 & 0.5357 & 0.7737 \\
\textbf{Ours} & \textbf{284.92} & \textbf{0.8407} & \textbf{0.0654} & \textbf{0.9279} & 0.9468 & 0.9896 & \textbf{0.4701} & \textbf{0.5428} & \textbf{0.7754} \\
\bottomrule
\end{tabular}}
\end{table}

\begin{table}[t]\centering
\caption{Component ablation on both benchmarks. Each row adds one component to the base student distilled with Self-Forcing (SF): AS denotes the attention sink, PR the pose-retrieved loop-closure memory (Sec.~\ref{sec:method_memory}), and GB the geometry-guided attention bias (Sec.~\ref{sec:method_bias}).}
\label{tab:ablation}
\begin{tabular}{lccc|ccc}
\toprule
& \multicolumn{3}{c|}{TartanGround-Revisit} & \multicolumn{3}{c}{TartanAir-Revisit} \\
\cmidrule(lr){2-4} \cmidrule(lr){5-7}
Configuration & Keypoint$\uparrow$ & DINO$\uparrow$ & L1$\downarrow$ & Keypoint$\uparrow$ & DINO$\uparrow$ & L1$\downarrow$ \\
\midrule
SF & 23.26 & 0.6087 & 0.1148 & 195.65 & 0.7623 & 0.0862 \\
SF + AS & 39.40 & 0.6761 & 0.1078 & 257.05 & 0.8110 & 0.0721 \\
SF + AS + PR & 47.84 & \textbf{0.6926} & \textbf{0.1052} & 283.30 & \textbf{0.8410} & 0.0655 \\
SF + AS + PR + GB (\textbf{Ours}) & \textbf{49.08} & 0.6904 & \textbf{0.1052} & \textbf{284.92} & 0.8407 & \textbf{0.0654} \\
\bottomrule
\end{tabular}
\end{table}

\begin{figure}[htb!]
    \centering
    \includegraphics[width=\linewidth]{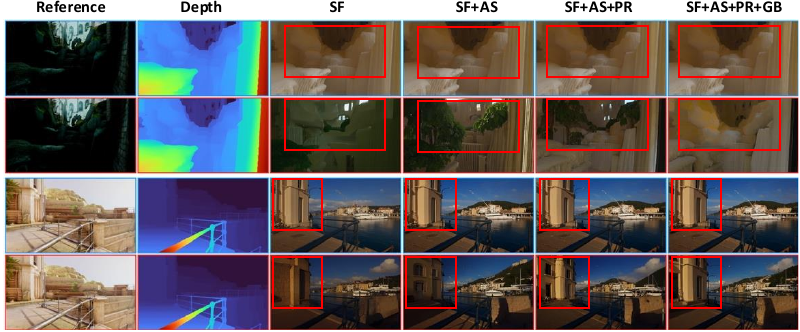}
    \caption{\textbf{Qualitative ablation.} Columns cumulatively add our components to Self-Forcing (SF): attention sink (AS), pose-retrieved memory (PR), and geometry-guided attention bias (GB). Rows pair the first visit (blue) with the return (red); red boxes mark the same region.}
\label{fig:ablation}
\vspace{-4mm}
\end{figure}

\subsection{Quantitative and Qualitative Results}
Tables~\ref{tab:tg-revisit} and~\ref{tab:ta-revisit} report results on TartanGround-Revisit and TartanAir-Revisit. Every training-free cache-management scheme improves revisit consistency over the base model, confirming that memory management is the right lever for this problem; our method pushes further by selecting memory with camera geometry and achieves the best score on every consistency metric, roughly doubling the base model's high-confidence keypoint matches on TartanGround-Revisit. And the same ordering holds on TartanAir-Revisit. On video quality, our method attains the best VBench-Long mean on both benchmarks, demonstrating that consistency gains of our method do not cost visual fidelity. Qualitatively (Fig.~\ref{fig:qual_compare}), baselines may regenerate altered content on return such as changed facades and inserted structures, while our method re-renders the boxed regions consistently across the loop.

\subsection{Ablation Study}
Table~\ref{tab:ablation} adds our components one at a time to the Self Forcing (SF) baseline. The attention sink (AS) yields a large gain by stabilizing global appearance against drift. Pose retrieval (PR) is the dominant revisit-specific factor: access to the actual earlier view is what loop closure requires. The geometric bias (GB) further improves keypoint matches, it refines \emph{where} each token reads within the retrieved chunk, exactly the placement property the matcher-based metric measures. Fig.~\ref{fig:ablation} shows the progression qualitatively.

\subsection{Integration with Real Game Engine}
\begin{figure}[htb!]
    \centering
    \includegraphics[width=\linewidth]{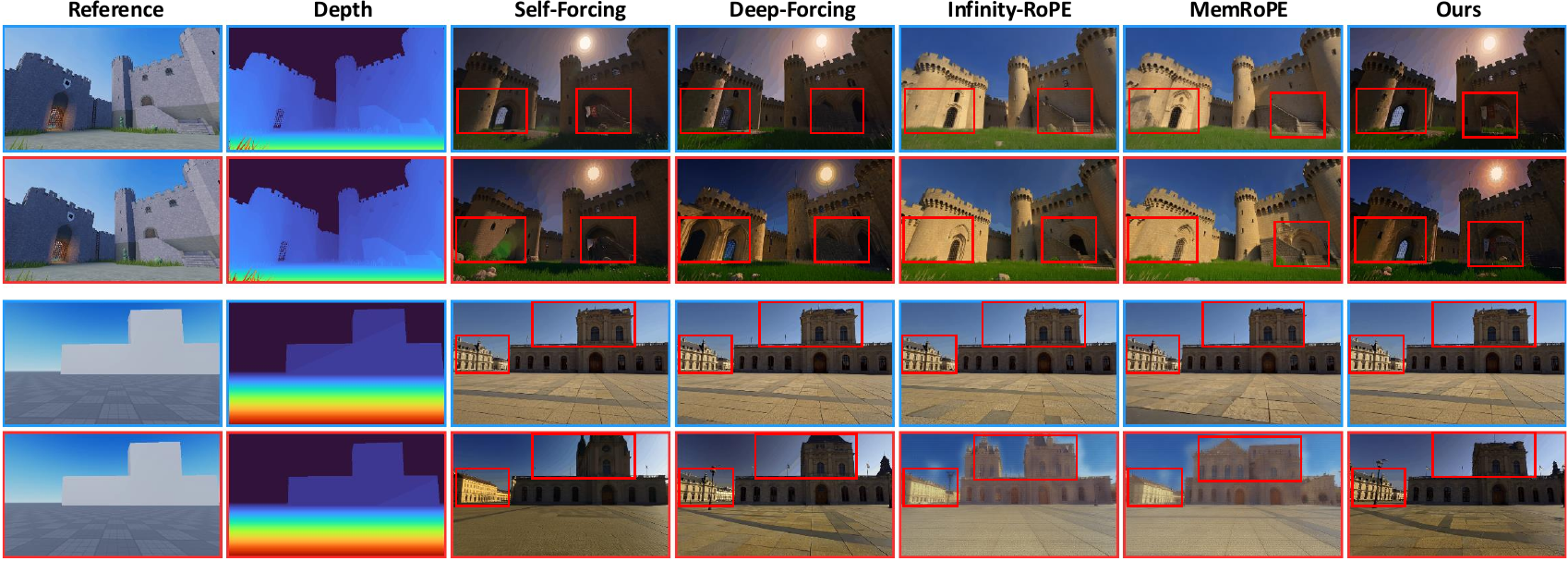}
    \caption{\textbf{Integration with an in-house game engine} on loop-closure paths (blue: first visit; red: return). On return, baselines alter structures; ours re-renders the same structures.}
\label{fig:engine}
\vspace{-3mm}
\end{figure}

We integrate our approach with an in-house game engine that streams exactly the annotations assumed in Sec.~\ref{sec:method_setup}: camera poses and metric depth, directly to the generator. Fig.~\ref{fig:engine} shows two representative cases on loop-closure camera paths: a castle with complex architecture (top) and a scene built from untextured geometric primitives (bottom). In both case, the generator must dress with plausible appearance. Compared with baseline methods that regenerate altered structures (e.g., an added dome), our method re-renders the same structures at visits. 

\section{Conclusion}
We presented a training-free approach to revisit-consistent streaming generative rendering. Exploiting correspondences a 3D engine already provides, our method retrieves pose-matched historical chunks into the bounded KV cache as loop-closure memory and steers attention toward geometrically corresponding tokens via a depth-reprojection Gaussian bias. The main limitation is the reliance on engine-supplied camera poses and metric depth; estimated geometry from online reconstruction methods like VGGT~\cite{wang2025vggt} and VGGT-$\Omega$~\cite{wang2026vggtomega} would extend the method to real video, at the cost of correspondence noise that our soft attention bias is designed to tolerate.

\bibliography{references}
\bibliographystyle{iclr2026_conference}

\appendix

\end{document}